\documentclass[conference]{IEEEtran}
\usepackage{blindtext}
\usepackage{etoolbox}
\makeatletter
\patchcmd{\@makecaption}
  {\\}
  {.\ }
  {}
  {}
\usepackage{cite}
\usepackage[colorlinks]{hyperref}
\ifCLASSINFOpdf
  \usepackage[pdftex]{graphicx}
\else
  \usepackage[dvips]{graphicx}
\fi
\usepackage[cmex10]{amsmath}
\usepackage{algorithm}
\usepackage{algpseudocode}
\usepackage{verbatim}
\usepackage{enumitem}
\usepackage{array,multirow}
\usepackage{color}
\usepackage{booktabs}
\usepackage{url}
\usepackage[normal]{threeparttable}

\newcommand{\figref}[1]{Fig.~\ref{#1}}

\newcommand{\secref}[1]{Section~\ref{#1}}
\newcommand{\tabref}[1]{Table~\ref{#1}}

\begin{document}
\title{Towards Efficient Convolutional Neural Network for Domain-Specific Applications on FPGA}
\author{
\IEEEauthorblockN{
    Ruizhe Zhao\IEEEauthorrefmark{1},
    Ho-Cheung Ng\IEEEauthorrefmark{1},
    Wayne Luk\IEEEauthorrefmark{1} and 
    Xinyu Niu\IEEEauthorrefmark{2}
}
\IEEEauthorblockA{
    \IEEEauthorrefmark{1}Department of Computing, Imperial College London, London, United Kingdom\\
    \textit{\{ruizhe.zhao15, h.ng16, w.luk\}@imperial.ac.uk}}
\IEEEauthorblockA{
    \IEEEauthorrefmark{2}Corerain Technologies Ltd., Shenzhen, China. \textit{xinyu.niu@corerain.com}}
}
\maketitle

\IEEEpeerreviewmaketitle
\begin{abstract}
FPGA becomes a popular technology for implementing Convolutional Neural Network (CNN) in recent years. Most CNN applications on FPGA are domain-specific, e.g., detecting objects from specific categories, in which commonly-used CNN models pre-trained on general datasets may not be efficient enough. This paper presents TuRF, an end-to-end  CNN acceleration framework to efficiently deploy
domain-specific applications on FPGA by transfer learning that adapts pre-trained models to specific domains, replacing standard convolution layers with efficient convolution blocks, and applying layer fusion to enhance hardware design performance. We evaluate TuRF by deploying a pre-trained VGG-16 model for a domain-specific image recognition task onto a Stratix V FPGA. Results show that designs generated by TuRF achieve better performance than prior methods for the original VGG-16 and ResNet-50 models, while for the optimised VGG-16
model TuRF designs are more accurate and easier to process.
\end{abstract}
\section{Introduction}\label{sec:intro}

There has been much recent work on developing FPGA implementations of \emph{Convolutional Neural Networks} (CNNs).
While significant progress has been made in optimising the inference process of general CNN models on FPGAs,
training and optimising CNNs for various \emph{domain-specific applications} remain a demanding task.
CNN models for domain-specific applications only need to detect or classify objects from a narrow range of classes.
Recent discovery in \emph{transfer learning}~\cite{Bengio2011a} ---
a research topic focusing on exploiting features reusable from one task to another ---
shows that CNN models that are pre-trained on general datasets
can be efficiently \emph{fine-tuned}~\cite{flowers} for specific domains.
This approach works well for medical image analysis:
a pre-trained CNN with adequate fine-tuning can outperform or perform as well as training from scratch~\cite{Tajbakhsh2017}.

While the transfer learning approach is promising,
the challenge is to exploit it for domain-specific applications on FPGA,
where efficient processing is vital.
For tasks in a specific domain,
standard convolution layers dedicated to extracting general features are over-parameterised and
can be replaced by efficient \emph{convolution blocks},
which consist of multiple small convolution layers with much fewer parameters.
Example blocks are
\emph{bottleneck}~\cite{He2015}, \emph{depthwise separable}~\cite{Howard2017},
and \emph{separable bottleneck}~\cite{Sandler2018}.
They can reduce computational redundancy while maintaining a satisfactory accuracy.
Meanwhile, since a layer-replaced model normally can be easily fine-tuned,
the cost of layer replacement is minor.
However,
they are rarely explored and implemented in any of the previous work on FPGA acceleration of CNNs.

\begin{figure}[!t]
\centering
\includegraphics[width=0.48\textwidth]{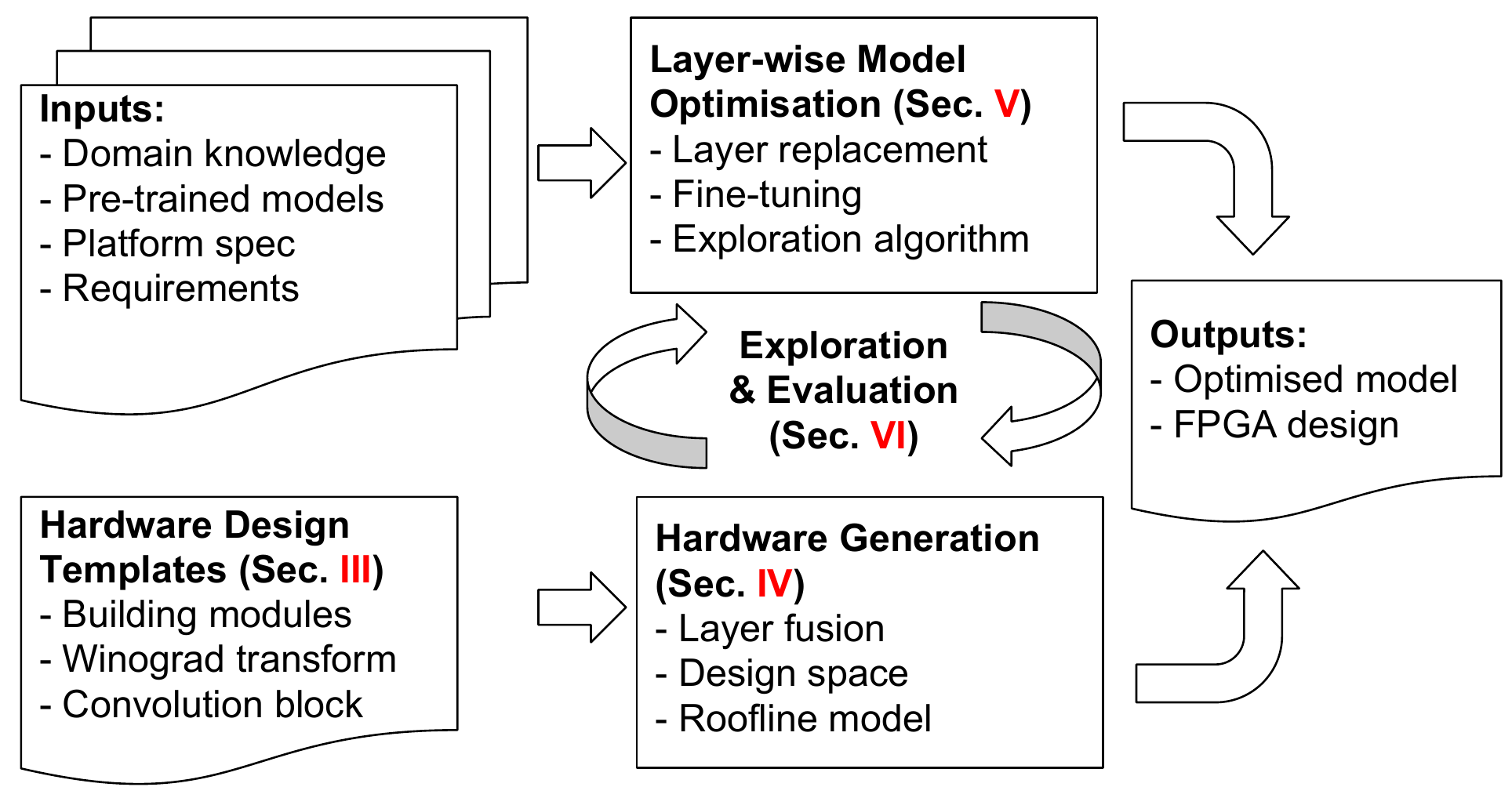}
\vspace{-1em}
\caption{TuRF design flow. Corresponding sections are {\color{red}marked}.}
\vspace{-1em}
\label{fig:diagram}
\end{figure}

This paper proposes \emph{TuRF},
a novel framework that generates efficient CNN models on FPGA for domain-specific applications (\figref{fig:diagram}).
TuRF accepts a CNN model pre-trained from a large-scale dataset,
replaces its selected standard convolution layers with various convolution blocks,
fine-tunes and evaluates the layer-replaced model,
and outputs an efficient FPGA design in the end.
To efficiently process convolution blocks,
TuRF generates FPGA designs that \emph{fuse} their inner convolution layers.
The major contributions are as follows:
\begin{enumerate}[leftmargin=*]
\item
A design template that supports efficient CNN models and convolution blocks
with Winograd~\cite{Lavin2015}, and also layer fusion optimisation
(Section~\ref{sec:design} and ~\ref{sec:design:fusion}).
\item
Characterisation of the design space of CNN model regarding domain-specific applications
and a transfer learning inspired layer-wise optimisation
that replaces standard convolution layers by blocks with fine-tuning (Section~\ref{sec:model}).

\item
Evaluation of our framework with flower classification~\cite{flowers},
a domain-specific application for transfer learning evaluation.
Results show that on Stratix V 5GSD8,
our framework can generate both efficient hardware and better CNN models
for a given application (Section~\ref{sec:eval}).

\end{enumerate}



\section{Motivations and Background}\label{sec:background}

Implementing CNN onto FPGA for domain-specific applications is challenging.
Most of the previous efforts focus on data quantisation and binarisation~\cite{Umuroglu2017, Qiu2016},
arithmetic transformations~\cite{Lu2017}, or
exploiting model sparsity with pruning~\cite{Han2017}.
However, 
it is difficult to apply these methods from a domain-specific perspective because
their evaluations and results are based on
specific CNN models, and are not guaranteed to be reproducible using other models.
Also, no clear correlation between accuracy
and data representation or sparsity has been discovered yet.
Finally, a sparse CNN model is much harder to process on FPGA
and different sparsity patterns can result in very different performance.

\begin{figure}[!t]
\centering
\includegraphics[width=0.48\textwidth]{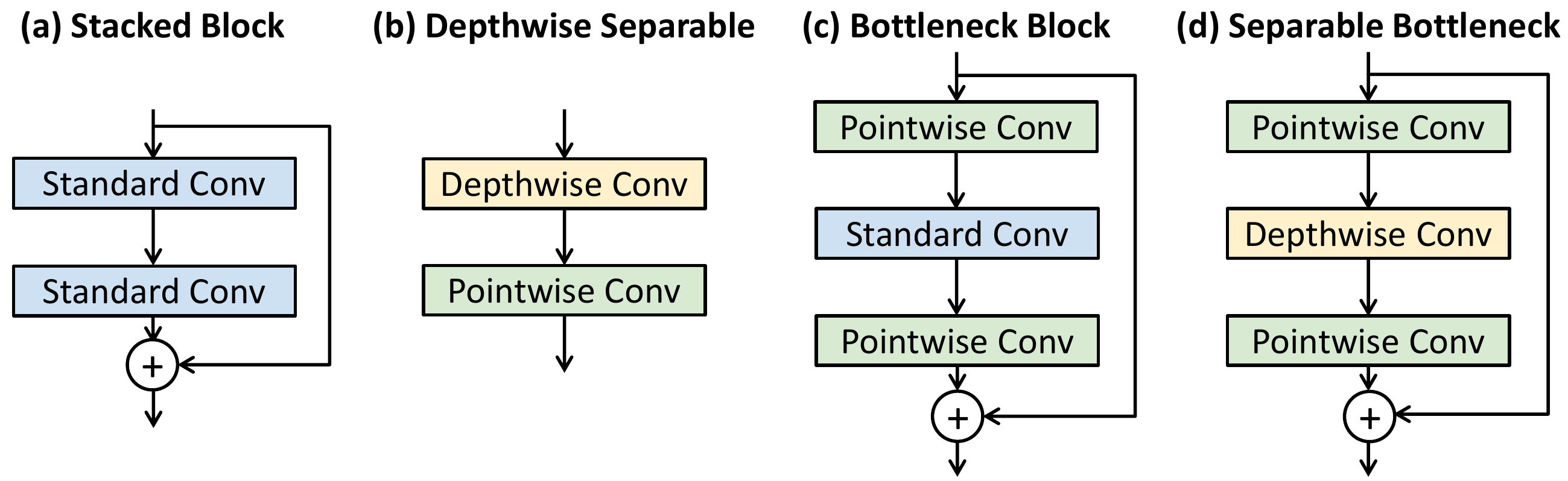}
\vspace{-1em}
\caption{Efficient convolution blocks that aim at improving the efficiency.
}
\label{fig:conv_blocks}
\vspace{-1em}
\end{figure}

The motivation of the proposed framework rests on the recent trend
in efficient CNN architecture design~\cite{Howard2017, Sandler2018, He2015}.
Essentially, the redundancy in CNN is sometimes architectural and can be reduced by modifying
the standard convolution layer with efficient \emph{convolution blocks}
as shown in \figref{fig:conv_blocks}. 
A convolution block is a set of convolution layers that extracts features as a whole.
More importantly,
a convolution block generally consumes fewer resources than its equivalent standard convolution layer.
Compared to quantisation and pruning which rely on statistical properties of pre-trained CNN models,
this approach is more generic and has been evaluated
for different domain-specific applications, as shown in~\cite{Howard2017}.
It is one of the major tasks for TuRF to explore the optimisation opportunity of switching from
convolution layers to blocks.
Nevertheless, there is no dedicated FPGA implementation for convolution blocks,
and therefore,
TuRF also aims at accelerating CNN for domain-specific applications
by exploring the FPGA implementation of convolution blocks.

\subsection{Convolution Layer and Convolution Block}
A convolution layer correlates an input feature map $\mathbf{D}$ and a filter $\mathbf{G}$ together.
Suppose $\mathbf{D}$ is an image with $C$ channels and spatial dimensions $H\times W$,
and $\mathbf{G}$ is a 4D filter that consists of $F$ output channels, $C$ input channels, and kernels of size $K\times K$,
the resulting feature map $\mathbf{Y}$ is defined as~\eqref{eq:conv},
where $*$ is the spatial convolution operator.
\vspace{-0.8em}
\begin{equation}\label{eq:conv}
\begin{footnotesize}
\begin{aligned}
\mathbf{Y}_{f}     &= \sum_{c=1}^{C} \mathbf{D}_{c} * \mathbf{G}_{f,c} \\
\mathbf{Y}_{f,x,y} &= \sum_{c=1}^{C} \sum_{h=1}^{K} \sum_{w=1}^{K} 
                      \mathbf{D}_{c,x+h,y+w} \times \mathbf{G}_{f, c, h, w}
\end{aligned}
\end{footnotesize}
\vspace{-0.5em}
\end{equation}

A standard convolution layer can be replaced by \emph{convolution block} to
improve efficiency. There are basically four types:

\subsubsection{Stacked Block}
It simply stacks two standard convolution layers together and reduces the number of channels.
Its input and output are connected by a \emph{shortcut} connection.
Please refer to the model ResNet-34~\cite{He2015} for more details.

\subsubsection{Depthwise Separable}
It is proposed in~\cite{Garcia2012, Chollet2016, Howard2017},
where the \emph{spatial} and \emph{cross-channel} correlation is studied separately
using \emph{depthwise} and \emph{pointwise} convolution respectively.
The depthwise convolution only performs spatial convolution in
each channel of the input feature map,
and the pointwise convolution is a special case of the standard convolution
by setting $K$ to 1.
Assume that $\mathbf{\widehat{G}}$ is the 3D depthwise filter
and $\mathbf{G}$ is the 2D pointwise filter,
\eqref{eq:dws_conv} defines the depthwise separable convolution layer
as described in~\cite{Howard2017}.
\vspace{-0.8em}
\begin{equation}\label{eq:dws_conv}
\begin{footnotesize}
\begin{aligned}
\mathbf{Y}_{f, x, y} = \sum_{c = 1}^{C} \left(\mathbf{D}_{c} * \mathbf{\widehat{G}}_{c}\right) \times \mathbf{G}_{f, c} 
\end{aligned}
\end{footnotesize}
\end{equation}

\subsubsection{Bottleneck Block}
A recent trend of constructing CNN is the prevailing use of \emph{bottleneck} block
as demonstrated in ResNet-50~\cite{He2015} which is 
economical and easy-to-train for deeper networks.
A bottleneck block consists of a stack of $1\times 1$, $3\times 3$, and $1\times 1$ convolution layers,
in which the first and last one reduce and increase the number of channels respectively.
The input of the bottleneck block is connected to the output of the previous stack where there also exists a residual connection
performing element-wise addition.

\subsubsection{Separable Bottleneck}
Evolved from the original bottleneck block,
\emph{linear bottleneck} is proposed and used in MobileNet V2~\cite{Sandler2018}
for efficiency improvement.
Compared to the original, the middle bottleneck convolution is replaced with its depthwise version
and the activation layer is removed after the last convolution,
so as to combine the efficiency of depthwise separable with the
effectiveness of bottleneck block.

\subsection{Winograd Algorithm}

As an arithmetic optimisation method dated back to 1980s,
the Winograd's minimal filtering algorithm~\cite{Winograd1980} is still proven to be powerful in optimising
convolution layer processing based on recent research~\cite{Lavin2015}.
An instance of 2D Winograd algorithm,
denoted by $F(m\times m, r\times r)$ where $m$ is the 2D tile size and $r$ is the filter kernel size, 
can be formulated as~\eqref{eq:winograd},
\vspace{-0.5em}
\begin{equation}\label{eq:winograd}
\begin{footnotesize}
\begin{aligned}
\mathbf{Y} =
\mathbf{A}^\mathsf{T}
 \Big[
 \left[ \mathbf{G}g\mathbf{G}^\mathsf{T} \right]
 \odot
 \left[ \mathbf{B}^\mathsf{T}d\mathbf{B} \right]
 \Big]
\mathbf{A}
= \mathbf{A}^\mathsf{T} \mathbf{X} \mathbf{A}
\end{aligned}
\end{footnotesize}
\vspace{-0.5em}
\end{equation}
where $\odot$ is the Hadamard product.
$\mathbf{G}$, $\mathbf{B}$, $\mathbf{A}$ are three transformation matrices
with $(m+r-1)\times r$, $(m+r-1)\times(m+r-1)$, $(m+r-1)\times m$ in shape.
$g$ is an $r^2$ filter kernel and $d$ with the size $(m+r-1)\times (m+r-1)$ is a tile of the input feature map. 
A widely used configuration is $F(4^2, 3^2)$, and compared to standard convolution with $4^23^2 = 144$ multiplications to produce $16$ output elements,
2D Winograd based convolution reduces the computation complexity to $6^2=36$ multiplications,
which is equivalent to $144 / 36 = 4$x speed-up.
For details about the Winograd Algorithm, please refer to~\cite{Lavin2015}.

The utilisation of Winograd for CNN acceleration on FPGA is discussed in~\cite{Lu2017, Aydonat2017, Yu2017, Shen2018}.
$F(4^2, 3^2)$ is applied in~\cite{Lu2017, Aydonat2017} and $F(2^2, 3^2)$ in~\cite{Yu2017, Shen2018}.
$F(4^2, 3^2)$ can achieve higher speed-up, but it contains constant factors that are not $2^n$.
In the following discussion, 
we use $F(4^2, 3^2)$ but our approach can be configured to support $F(2^2, 3^2)$.

\subsection{Efficient CNN Models}

\begin{figure}[!t]
\centering
\includegraphics[width=0.48\textwidth]{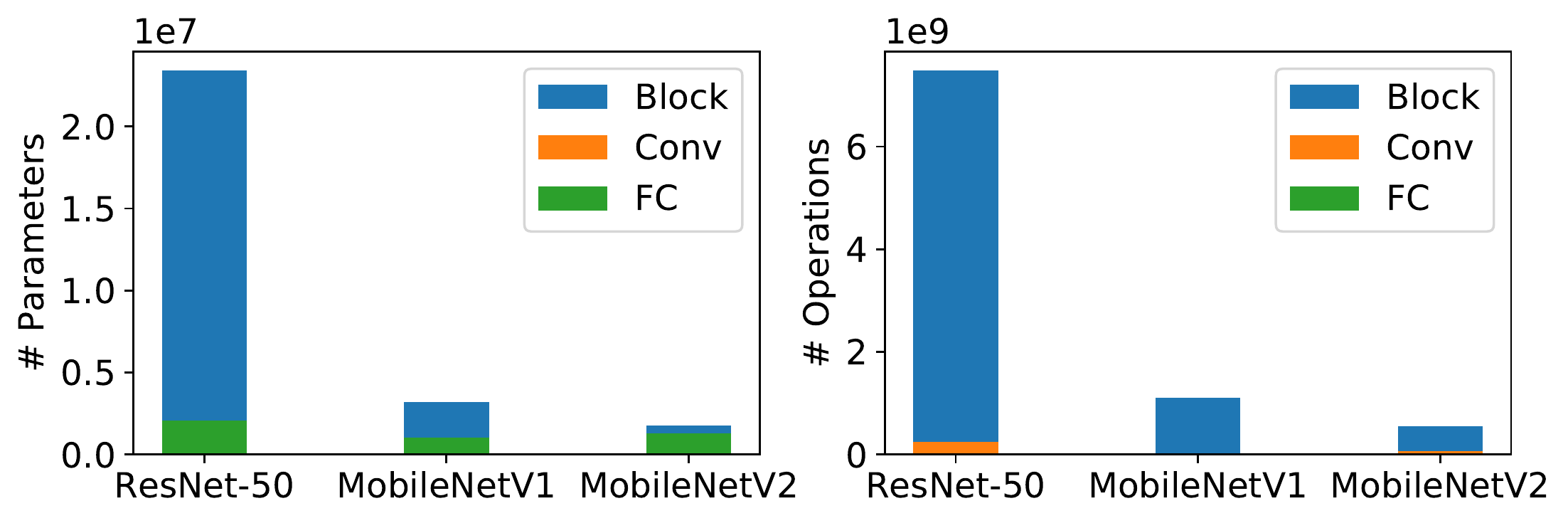}
\vspace{-1em}
\caption{A comparison of the number of parameters (left) and operations (right) within the convolution block, single convolution layer and FC layer between three efficient CNN models. Convolution blocks are dominating.}
\label{fig:net_stat}
\end{figure}

In this paper, we mainly study three efficient CNN models:
ResNet-50~\cite{He2015}, MobileNet V1~\cite{Howard2017} and V2~\cite{Sandler2018}, to gain research insight for our hardware template and to make a comparison to our generated CNN model for domain-specific application.
Default configurations are used for these networks.
As shown in~\figref{fig:net_stat}, convolution blocks are responsible for most of the operations and parameters.
These models are also compared to VGG-16~\cite{Simonyan2015a} as shown in \tabref{tab:net_stat}.
ImageNet top-1 accuracy of each model is listed in the same table as well.

\begin{table}[!t]
\centering
\caption{CNN Models Statistics.}
\renewcommand{\arraystretch}{1.1}
\label{tab:net_stat}
\begin{threeparttable}
\vspace{-0.8em}

\begin{tabular}{|c|c|c|c|c|}
\hline
\textbf{Model} &
\textbf{Block\tnote{1}} &
\textbf{Ops (GOPS)} &
\textbf{Params (M)} &
\textbf{Top-1 (\%)}\\\hline
\textbf{VGG-16} & --- & 30.95 & 138.3 & 71.5 \\\hline 
\textbf{ResNet-50} & (c) & 7.72 & 24.3 & 75.2 \\\hline 
\textbf{MobileNetV1} & (b) & 1.14 & 4.01 & 70.9\\\hline 
\textbf{MobileNetV2} & (d) & 0.61 & 3.31 & 71.9\\\hline 
\end{tabular}
\begin{tablenotes}
\item[1] (b), (c), (d) are symbols that denote convolution blocks in~\figref{fig:conv_blocks}.
\end{tablenotes}
\end{threeparttable}
\vspace{-1em}
\end{table}

\section{Hardware Design Template}\label{sec:design}

TuRF proposes the use of a scalable hardware design template as a fundamental component
in our framework.
The template enables generation of optimised CNN hardware by supporting various convolution types.
Particularly, the convolutional blocks discussed in \secref{sec:background}
are the primary research focus in our template development process.
Winograd transformation is also used to accelerate spatial convolution.

\subsection{Design Template Overview}

Our design template can be configured to support all the layers utilised in recent efficient CNN models.
We focus on convolution layer and convolution blocks, which are the most time-consuming parts in these models.
An accelerator for convolution layer or block can be constructed by basic \emph{building modules} in our template.
Each module can be configured regarding the level of parallelism or computation sequence.
Our design is by default implemented with \emph{fixed-point} representation,
and its configuration is decided by the data range.

Similar to~\cite{Venieris2016}, a design module is described as a tuple
$\langle \mathit{cfg}, in, out \rangle$
in which $\mathit{cfg}$ is a set of module configuration
and $in$, $out$ specify the width of input and output streams respectively.
The module configurations can be described with:
\begin{enumerate}
\item \emph{Tile shape}: $T_h, T_w, T_c$ denote the height, width, and channels of the input
and $T_f$ represents the output channels.
\item \emph{Level of parallelism}:
$P_c, P_f, P_h, P_w$ represent the number of elements to process in parallel along the input and output channels, height, and width axis.
\item \emph{Layer specifics}: $K$ denotes the kernel size and $r$ mentioned in \secref{sec:background} is replaced by $K$.
\end{enumerate}

\subsection{Basic Building Modules}\label{sec:design:basic}

\subsubsection{Line Buffer}
Given by \eqref{eq:lbuf}, it is a module that creates \emph{sliding windows} over an input feature map.
We use $K^\prime$ to denote either convolution kernel size $K$ or
the Winograd input tile size $(m + K - 1)$.
This module is implemented using \emph{shift registers} organised into $K^\prime$ rows.
\vspace{-0.5em}
\begin{equation}\label{eq:lbuf}
\begin{aligned}
\langle \{P_c, P_h, P_w\}, P_c P_h P_w, (K^\prime+P_h-1) (K^\prime+P_w-1) \rangle
\end{aligned}
\end{equation}
\vspace{-1.5em}

\subsubsection{Input and Output Buffers}
Buffers are implemented as on-chip memory to exploit the locality of the computation.
An input buffer caches input feature map to be reused throughout the computation,
and an output buffer stores and accumulates temporary results.
See \eqref{eq:ibuf} and~\eqref{eq:obuf} for descriptions.
%
\begin{align}
&\langle
    \{P_c, P_w\}, P_c\times P_w, P_c\times P_w
\rangle\label{eq:ibuf}\\
&\langle
    \{P_f, P_w\}, P_f\times P_w, P_f\times P_w
\rangle\label{eq:obuf}
\end{align}
\vspace{-1.5em}

\subsubsection{Winograd Transformation}
The Winograd algorithm is applied to standard and depthwise convolution to reduce the computation complexity.
According to~\eqref{eq:winograd},
three transformation modules are required to process a Winograd convolution.
Let $T_k$ be the Winograd tile size $(m + K - 1)$,
\eqref{eq:trans_ifmap}, \eqref{eq:trans_weights}, \eqref{eq:trans_ofmap}
illustrate the configurations and interfaces of the transformation
modules for input feature map $\mathbf{B}^\mathsf{T}d\mathbf{B}$,
weights $\mathbf{G}d\mathbf{G}^\mathsf{T}$,
and output $\mathbf{A}^\mathsf{T}\mathbf{X}\mathbf{A}$ respectively.
Each transformation consists of two multiplications between an input and a constant matrix,
which are implemented with either multipliers with LUTs 
or shift operators (for $2^n$ constants) to save resources.
%
\begin{align}
&\langle
\{P_c, K\}, P_c T_k^2, P_c T_k^2 
\rangle\label{eq:trans_ifmap}\\
&\langle
\{P_c, P_f, K\}, P_c P_f K^2, P_c P_f T_k^2
\rangle\label{eq:trans_weights}\\
&\langle
\{P_c, P_f, K\}, P_c P_f T_k^2, P_c P_f m^2
\rangle\label{eq:trans_ofmap}
\end{align}
\vspace{-1.5em}


\subsubsection{Arithmetic Module}
Most of the arithmetic computations in a typical CNN workload is dot-product,
which is employed in the spatial and cross-channel convolution and also in the fully-connected (FC) layers.
Each dot-product module consists of an array of multipliers followed by an adder tree.
The dot-product modules are further organised into a higher-level array for parallelisation.
This module can be shared among convolution and FC layers when necessary.

%

\subsubsection{Other Design Modules}
An \emph{element-wise addition} module performs addition of two identically sized feature maps.
An \emph{activation} module implements non-linear activation functions.
A \emph{normalisation} module normalises its input by Batch Normalisation.
We omit details about these modules 
because they are simple and have limited impact on the overall performance.

\subsection{Implementation of a Single Layer}\label{sec:design:single}

\begin{figure}[!t]
\centering
\includegraphics[width=0.48\textwidth]{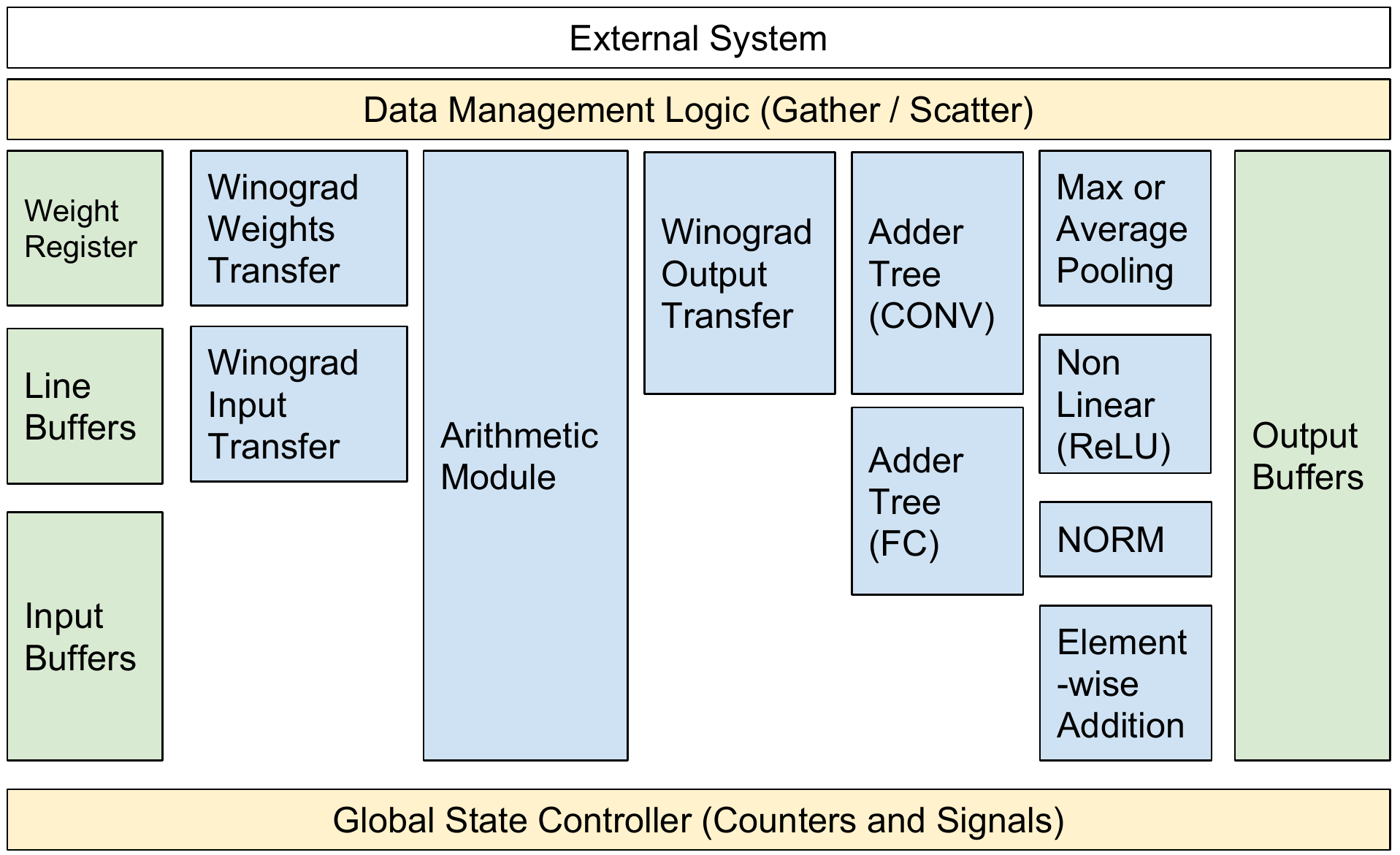}
\caption{Diagram of our proposed configurable system architecture
(NORM stands for normalisation).}
\label{fig:conv_arch}
\end{figure}

An accelerator can be constructed from building modules to perform the computation,
which can only perform the computation of one layer at one time,
in contrast to the fused layer design (\secref{sec:design:fusion}).
\figref{fig:conv_arch} shows the system diagram when a single layer is implemented.
It can support different types of convolution such as depthwise or pointwise and fully-connected layers by efficiently sharing
dot-products in the arithmetic module.
Modules are connected using data-flow streams with the same input and output width.
Outputs from building modules should be consumed immediately to avoid congestion.
A global state controller together with counters are utilised for each design
to assign addresses for buffers and to enable/disable read/write actions
and control data-flow directions.
Multiplexers are implicitly inserted between the columns in the figure to control computation.
The arithmetic module is shared by convolution with or without Winograd transformation,
and fully-connected layer.




The computation sequence of convolution,
which is a permutation of interchangeable nested loops indices $(f, c, i)$ based on~\eqref{eq:conv},
has a large impact on the architectural structure.
The impact of such permutation for individual convolution layer
has been extensively studied in recent research studies~\cite{Zhang2015, Ma2017}.
Therefore, we only discuss two computation sequences,
\emph{filter-major} $(f, c, i)$ and \emph{channel-major} $(c, f, i)$,
and present their impacts on buffer sizes and pipeline in the rest of this paper.
The size of the buffer for the major index is linear to its parallel factor.
For example, the output buffer, which is iterated with the $f$ index, is of size $P_f H W$ and is linear to $P_f$.
The pipeline behaviour between two adjacent layers can be different if their computation sequences are
configured in different ways (\ref{fig:seq_pipe}).
Further discussions in the next section.



\section{Fused Convolution Blocks}\label{sec:design:fusion}
%

Convolution blocks consume most of the operations according to~\figref{fig:net_stat}
and should be well-optimised for performance improvement according to Amdahl's law.
Similar to the previous work~\cite{Ma2017a},
a baseline accelerator for the convolution block is mainly based on a layer-by-layer execution.
This approach incurs significant off-chip data transfer and
consequently cannot fully exploit the potential of pipelining CNN layers.

\begin{figure}[!t]
\centering
\includegraphics[width=0.40\textwidth]{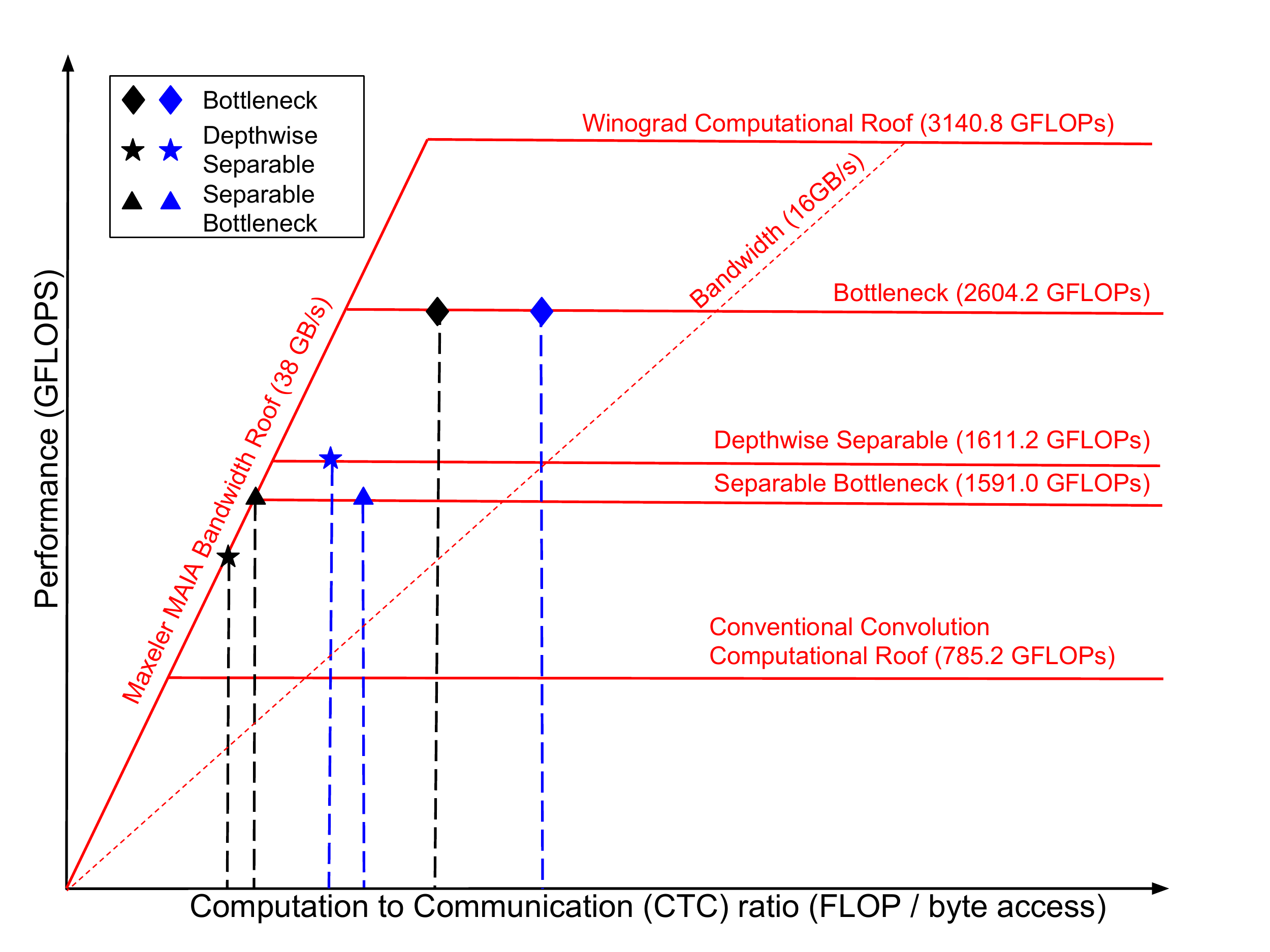}
\vspace{-1em}
\caption{We use the roofline model to evaluate the benefits of layer fusion for different convolution blocks.
The baseline is marked in black colour and the fused design is marked {\color{blue}blue} colour.
The target FPGA is Stratix-V 5SGSD8 on Maxeler MPC-X node.}
\label{fig:fused_roofline}
\end{figure}

To overcome this drawback,
we propose a fused accelerator for the convolution block
that enables the computation of all layers to complete in one launch.
The benefits of layer fusion are explained by the \emph{roofline model}~\cite{Williams2009} and
\figref{fig:fused_roofline} illustrates the possible benefits from layer fusion in different
convolution blocks.
We extend the analysis from~\cite{Lu2017} by adding computational roofs for three convolution blocks.
We have to note that the bandwidth of the evaluation system (Maxeler MPC-X Node) is so large that only depthwise separable blocks can take advantage of layer fusion. However, all convolution blocks can benefit from layer fusion if the bandwidth is decreased to $16\ GB/s$ or smaller which is common for commonality FPGA devices.

The idea of layer fusion is inspired by~\cite{Manoj2016, Xiao2017}.
These works only fuse the standard convolution with uniform kernel size and target high-end FPGAs.
Yet layer fusion is more difficult in our case
because there are other convolution variants,
and FPGAs do not always have sufficient resources to fully place the fused block.
Basically, we improve the previous approaches to address the following \emph{new} challenges:
\begin{enumerate}[leftmargin=*]
\item
The fusion method can support various convolution types.
\item
There are many options to explore when the layers are fused, such as buffer size and computation sequence.
\item
Tiling must be considered to support small FPGAs.
\end{enumerate}



\subsection{Fusion Method for General Layer}


We propose a method that can generate the fused design for typical convolution blocks automatically using the following steps.
1) The hardware implementation of convolution layer is selected layer by layer.
2) The input and output buffers between adjacent layers are combined.
3) The final configurations are aggregated and decided by the predicted latency. 

The first step can be easily implemented 
because our design template can support convolution types in any known blocks.
In the second step,
we decide the buffer usage between two adjacent layers by
the layer types and performance requirements.
If two adjacent layers are standard and pointwise convolution,
the use of a single buffer can minimise the area cost but may incur stalling of the entire pipeline, as the previous layer can only write into the buffer when the subsequent layer finishes the computation. Doubling the buffer can eliminate this issue with the increase of the area cost.

The third step determines the final configurations of the fused accelerator which includes:
the level of parallelism,
buffer sizes,
and computation sequences.
Suppose there are $N$ layers in a given module.
Let $\langle P_{h}^i, P_{w}^i, P_{c}^i, P_{f}^i \rangle$ be the parallelisation parameters of layer $i$.
To represent the computation sequences,
we use $Seq^i \in \{\mathtt{FM}, \mathtt{CM}\}$ to denote whether layer $i$
is filter-major $\mathtt{FM}$ or channel-major $\mathtt{CM}$.

\subsubsection{Parallelisation Parameters}
The parameters of a fused design should satisfy constraints in~\eqref{eq:fuse_param},
which ensures the widths between all the input and output ports along the design modules are the same.
Derived from~\eqref{eq:fuse_param}, parallelisation parameters of a fused design can be simplified as
$\langle P_h, P_w, P_c^1, P_c^2, \dots, P_c^N, P_f\rangle$.
\vspace{-1em}

\begin{equation}\label{eq:fuse_param}
\forall i \in \{2, \dots, N\}\ P_h^i = P_h^{i-1} \wedge P_w^i = P_w^{i-1} \wedge P_c^i = P_f^{i-1}
\end{equation}



\begin{table}[!t]
\renewcommand{\arraystretch}{1.3}
\caption{Buffer sizes under different configurations.}
\label{tab:buffer_sizes}
\centering

\vspace{-1em}
\begin{tabular}{|c|c|c|c|}
\hline
\textbf{$(Seq^{i-1}, Seq^i)$} &
\textbf{$L_{i-1}$ Output} & \textbf{$L_i$ Input} & \textbf{Double Buffering} \\
\hline
$\mathtt{(FM,CM)}$ & $P_c^i T_h^i T_w^i$ & $T_c^i T_h^i T_w^i$ & $2\times P_c^i T_h^i T_w^i$\\\hline
$\mathtt{(CM,FM)}$ & $T_c^i T_h^i T_w^i$ & $T_c^i T_h^i T_w^i$ & $2\times T_c^i T_h^i T_w^i$\\\hline
$\mathtt{(FM,FM)}$ & $P_c^i T_h^i T_w^i$ & $T_c^i T_h^i T_w^i$ & $2\times P_c^i T_h^i T_w^i$\\\hline
$\mathtt{(CM,CM)}$ & $T_c^i T_h^i T_w^i$ & inefficient         & $2\times T_c^i T_h^i T_w^i$\\\hline
\end{tabular}
\vspace{-1em}
\end{table}

\subsubsection{Buffer Size} 
The size of a buffer depends on the sequence of layers that it connects to.
The first input and the last output buffers are similar to the ones in \secref{sec:design:basic},
while other intermediate buffers are more complicated to analyse.
The size of buffer $B_i$, which is connected to layer $L_{i-1}$ and $L_i$,
depends on $Seq^{i-1}$, $Seq^i$, and the following options:
1) the same size as the buffer in $L_i$ input or $L_{i-1}$ output; 
2) double buffering to avoid stalling of the pipeline.

\tabref{tab:buffer_sizes} lists the size of the intermediate buffer $B_i$ under different configurations. 
The first column is the sequence of the two layers connected using a buffer.
Double buffering is only applied when it is indeed beneficial
for improving the pipeline performance.
A configuration is \emph{inefficient} if the buffer is too small to store
the required input or output.

\subsubsection{Computation Sequence and Pipeline}

\begin{figure}[!t]
\centering
\includegraphics[width=0.48\textwidth]{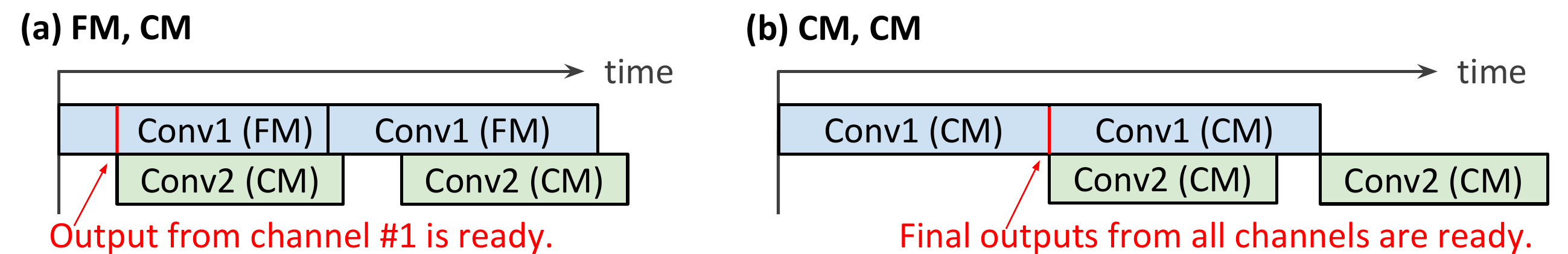}

\caption{Two example pipelines of a fused design with different sequences for a stacked block:
$\langle\mathtt{FM}, \mathtt{CM}\rangle$ and $\langle\mathtt{CM}, \mathtt{CM}\rangle$. Each rectangler box represents a complete execution of a tile.
Double buffering is applied in (a).}
\label{fig:seq_pipe}
\end{figure}

\begin{figure}[!t]
\centering
\includegraphics[width=0.48\textwidth]{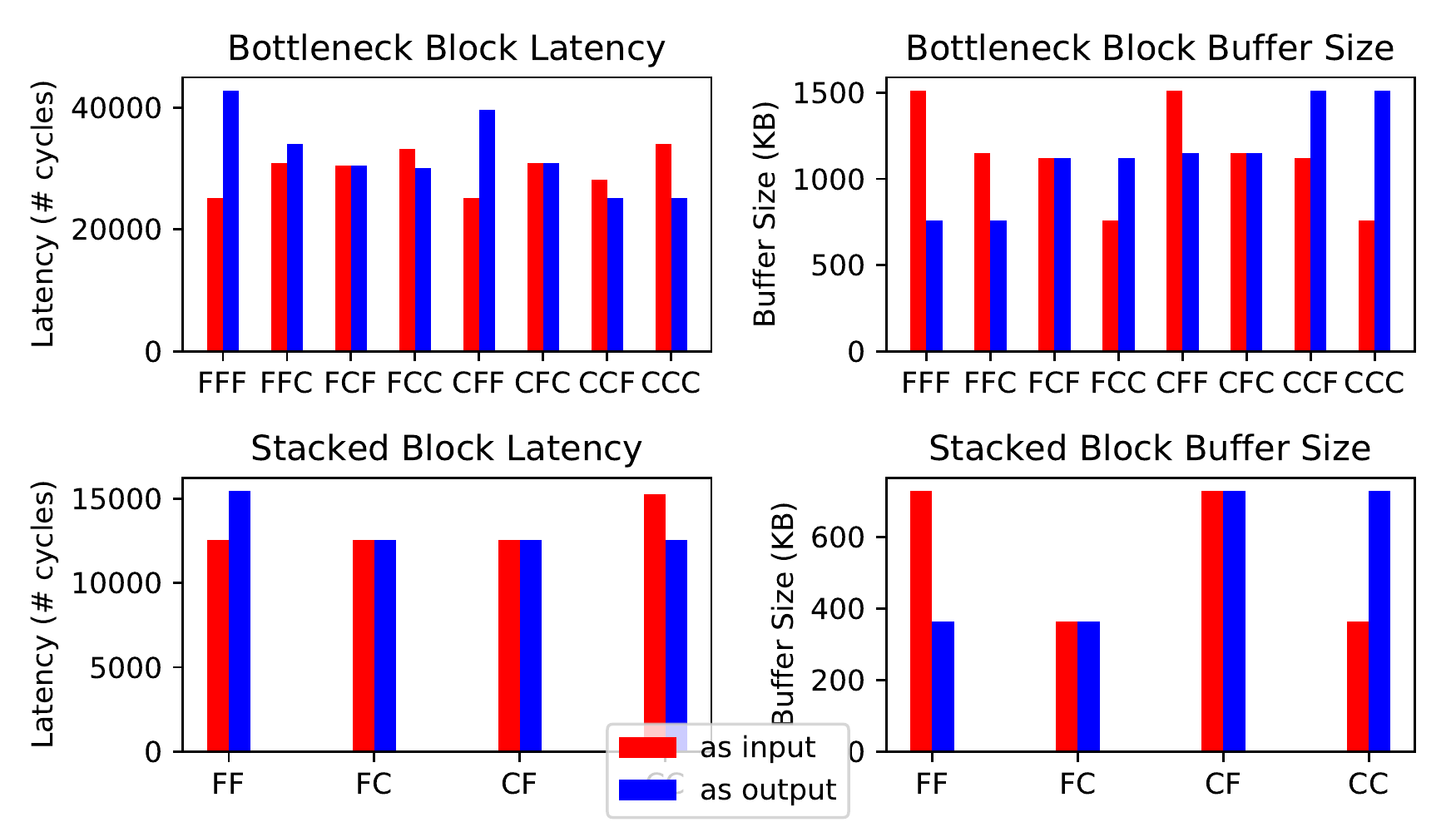}
\vspace{-1em}
\caption{Comparison of \emph{pipeline latency} and \emph{buffer size} with different computation sequences (x-axis)
and buffer options (bar colouring). In the x-axis, 'F' refers to filter and 'C' refers to channel, and hence 'FFF' means that every layer within the bottleneck block uses a filter-major sequence.
Moreover,
A {\color{red}red bar} refers to the adaption of the buffer size used in the previous layer while a {\color{blue}blue bar} refers to that used in the next layer.}
\label{fig:fusion_pipeline}
\end{figure}

The fused design is a streaming architecture and the computation of all layers are pipelined.
We notice that computation sequence of each layer can affect the pipelining
as illustrated in~\figref{fig:seq_pipe}.
$\langle\mathtt{FM}, \mathtt{CM}\rangle$ has a lower latency than $\langle\mathtt{CM}, \mathtt{CM}\rangle$
(used in~\cite{Xiao2017})
since the first output finalised by layer $1$ can be immediately consumed by layer $2$.
For more complicated cases,
we implement a cycle-accurate simulator to obtain all combinations of computation sequences
and evaluate their latency.
Apart from latency,
we also consider buffer sizes since they are  affected by computation sequences as shown in Table~\ref{tab:buffer_sizes} and 
\figref{fig:fusion_pipeline} shows the exploration result for the bottleneck and  stacked blocks.

\subsubsection{Tiling}


A convolution block can be tiled into smaller pieces when the on-chip resources are limited.
Specifically,
for a convolution block with $N$ layers,
a \emph{tile} can be defined as $\langle T_h, T_w, T_c^1, \dots, T_c^N, T_f\rangle$.
Unlike tiling a convolution layer which mainly introduces an off-chip transfer overhead,
tiling a convolution block can incur much redundant computation as well.
Therefore, tiling configurations should be carefully explored to avoid such cases.

\subsection{Design Space Exploration}

\begin{figure}[!t]
\centering
\includegraphics[width=0.48\textwidth]{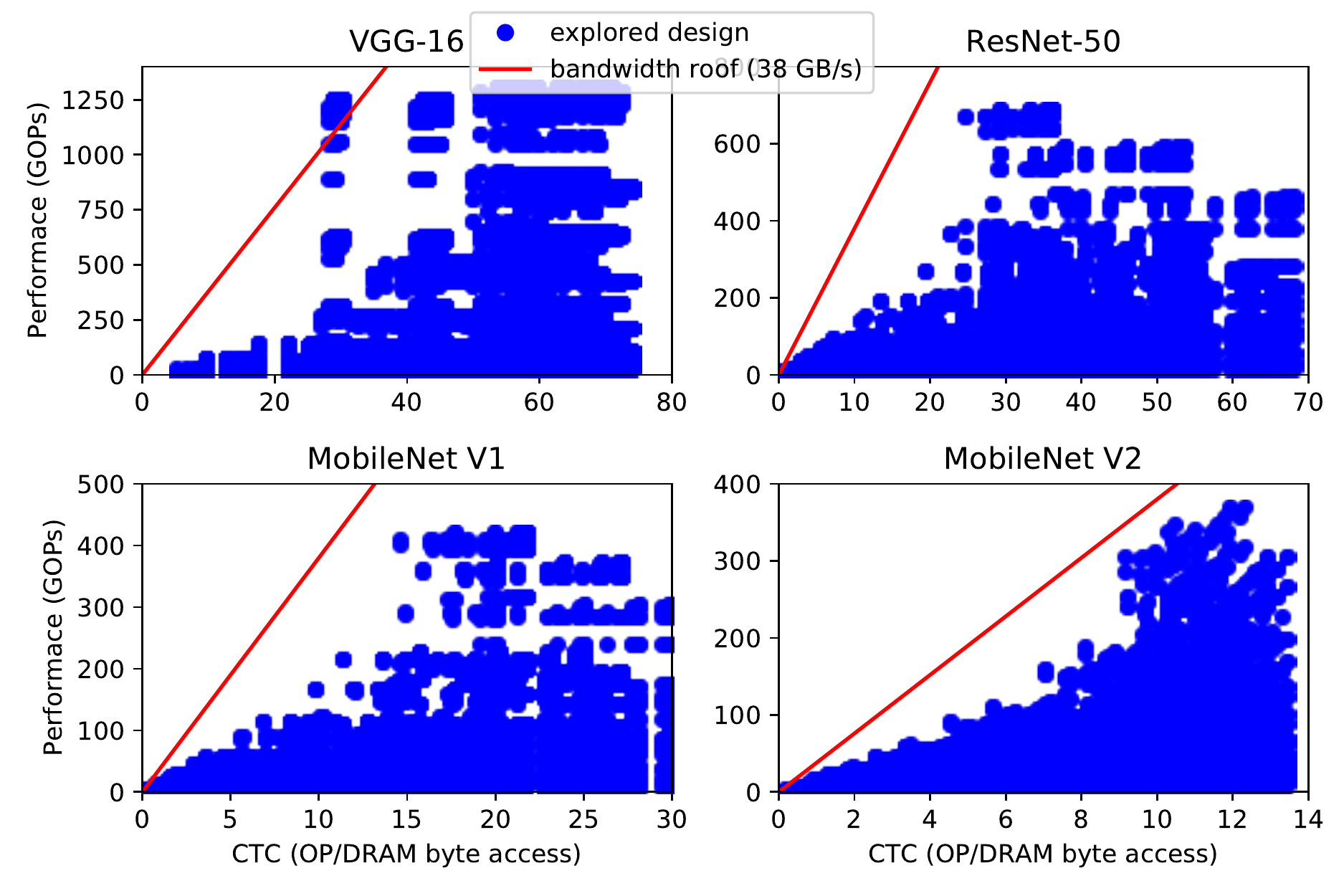}
\vspace{-1em}
\caption{Design space exploration of four CNN models targeting Stratix-V 5SGSD8 on Maxeler MPC-X node.
The computational roofs are not shown because they are above the upper limit of the $y$-axis.}
\label{fig:roofline}
\vspace{-1em}
\end{figure}

When the previously discussed configurations are combined,
we can characterise the hardware design space of a convolution block
by $\langle T_h, T_w, \{T_c^i\}, T_f, P_h, P_w, \{P_c^i\}, P_f, \{Seq^i\}\rangle$.
Winograd based design is used by default.
The performance and area cost are evaluated in two steps:
a cycle-accurate simulator is used to find the best computation sequence and its latency,
and the latency can provide the performance numbers;
and the resource consumption can be computed by a linear prediction model built upon synthesised designs.
Finally, the roofline model is used to find the best design under resource constraints.
\figref{fig:roofline} presents the exploration results for three efficient CNN models and a baseline VGG-16 model using our hardware template.
We notice that the performance in GOPs for the efficient models is generally smaller.
However, fewer operations are also required for these networks, and hence the overall inference time is still shorter (\secref{sec:eval}).


\begin{table*}[!t]
\centering
\begin{threeparttable}
\caption{Evaluation of Efficient CNN Models and Different VGG-16 Variants on Stratix V. }
\label{tab:eval_result}
\begin{tabular}{|l|l|l|l|l|l|l|l|l|}
\hline
\multirow{2}{*}{ }         & \multicolumn{3}{c|}{\textbf{Efficient CNN Models}}
                           & \multicolumn{4}{c|}{\textbf{VGG-16 Variants}\tnote{1}}
                           \\\cline{2-8}
                           & ResNet-50 & MobileNet V1  & MobileNet V2 & VGG-16  & VGG-16 (1) & VGG-16 (2) & VGG-16 (5)\\\hline
\textbf{\# Ops (GOP)      }& 7.74      & 1.14          & 0.611        & 30.95   & 26.29      & 19.36      & 3.82      \\\hline
\textbf{\# Param. (M)     }& 25.5      & 4.21          & 3.47         & 138.3   & 132.1      & 129.83     & 125.3     \\\hline
\textbf{Clock Freq. (MHz) }& 200       & 200           & 200          & 200     & 200        & 200        & 200       \\\hline
\textbf{Bit width         }& 16 bit    & 16 bit        & 16 bit       & 16 bit  & 16 bit     & 16 bit     & 16 bit    \\\hline
\textbf{DSP usage         }& 1680      & 1664          & 1856         & 1738    & 1872       & 1536       & 1680      \\\hline
\textbf{Latency (ms)      }& 7.95      & 0.884         & 1.02         & 14.5    & 10.3       & 9.65       & 8.42      \\\hline
\textbf{Throughput (GOPS) }& 973.2     & 1287.2        & 592          & 1928.4  & 2561.5     & 2007.0     & 453.6     \\\hline
\textbf{Top-1 Accuracy    }& 93.5      & 88.3          & 87.5         & 90.5    & 93.5       & 92.75      & 84.75     \\\hline
\end{tabular}

\begin{tablenotes}
\footnotesize
\item[1] The $n$ in VGG-16 ($n$) denotes a VGG-16 variant that has $n$ number of layers replaced by
depthwise separable convolution.
\end{tablenotes}
\end{threeparttable}
\end{table*}

\section{Layer-wise Model Optimisation}\label{sec:model}

The objective of TuRF is to find an efficient CNN model and its corresponding design on FPGA
for a given domain-specific application.
Section~\ref{sec:design} and~\ref{sec:design:fusion} discuss how to map efficient CNN models on FPGA designs.
Moreover, in this section,
we look into the design space exploration of CNN model,
which is inspired by transfer learning for layer-wise optimisation,
and the final tool-flow for TuRF.

\subsection{CNN Model Selection and Optimisation}
CNN model optimisation is about searching for the most efficient network under
pre-defined accuracy requirements.
If the hardware factor is put aside for now,
we can define model efficiency as the number of parameters and operations required
to achieve a certain accuracy,
and the remaining challenges are the characterisation and exploration of the model design space.

\subsubsection{Model Design Space}

A typical CNN model is a sequence of cascading layers with convolution layers. 
To restrict the scale of the design space,
we only explore models that are \emph{grown} from either VGG-16 or ResNet-50.
To further limit the design space for feasible exploration,
a model originated from VGG-16 or ResNet-50 can have
its convolution layers replaced only by a particular \emph{separable} convolution block as shown in~\eqref{eq:corr}.
Such replacements are also the partial motivation for the design of MobileNet.
We represent a model in our design space as
$\langle M, L^1, \dots, L^N \rangle$,
in which $M \in \{\text{VGG-16, ResNet-50}\}$ is the base model with $N$ convolution layers
and $L^i\in\{ \mathtt{ORIGIN}, \mathtt{SEPARABLE}\}$ indicates replacement.


\vspace{-1em}
\begin{equation}\label{eq:corr}
\begin{aligned}
\textit{standard convolution} &\longrightarrow \textit{depthwise separable block} \\
\textit{bottleneck block}     &\longrightarrow \textit{separable bottleneck block}
\end{aligned}    
\end{equation}


\subsubsection{Exploration Method}

In most cases we do not have a sufficient budget for training every possible model.
Therefore,
we devise the following optimisation approach, inspired by the principles of transfer learning.
The input to our exploration procedure can be any models pre-trained based on ImageNet,
which supposedly is general and consists of removable redundancies regarding the targeting application.
We intend to achieve the required accuracy by \emph{fine-tuning} the input model,
in which only top layers are trained and others are fixed.
We also assume that replacing top convolution layers are more beneficial than bottom ones.
This is based on~\cite{Zeiler2014} explaining
the mechanism of CNN for computer vision,
that convolution layers closer to the bottom extract lower-level features
such as edges and shapes,
and those closer to the top understand the higher-level features such as faces and eyes.
Hence, our assumption is mostly valid because the model is now focusing high-level domain knowledge.

As such, we propose a heuristic, greedy algorithm to explore model design space.
It starts with a pre-trained model and tries to replace layers from the top.
In each iteration, this algorithm fine-tunes the model candidate for a fixed number of steps.
The procedure terminates once the algorithm fails to satisfy the accuracy requirement.
Note that when the budget is sufficient, we need not stop and can continue searching.

\subsubsection{Evaluation}\label{sec:model:evaluation}

\begin{figure}[!t]
\centering
\includegraphics[width=0.48\textwidth]{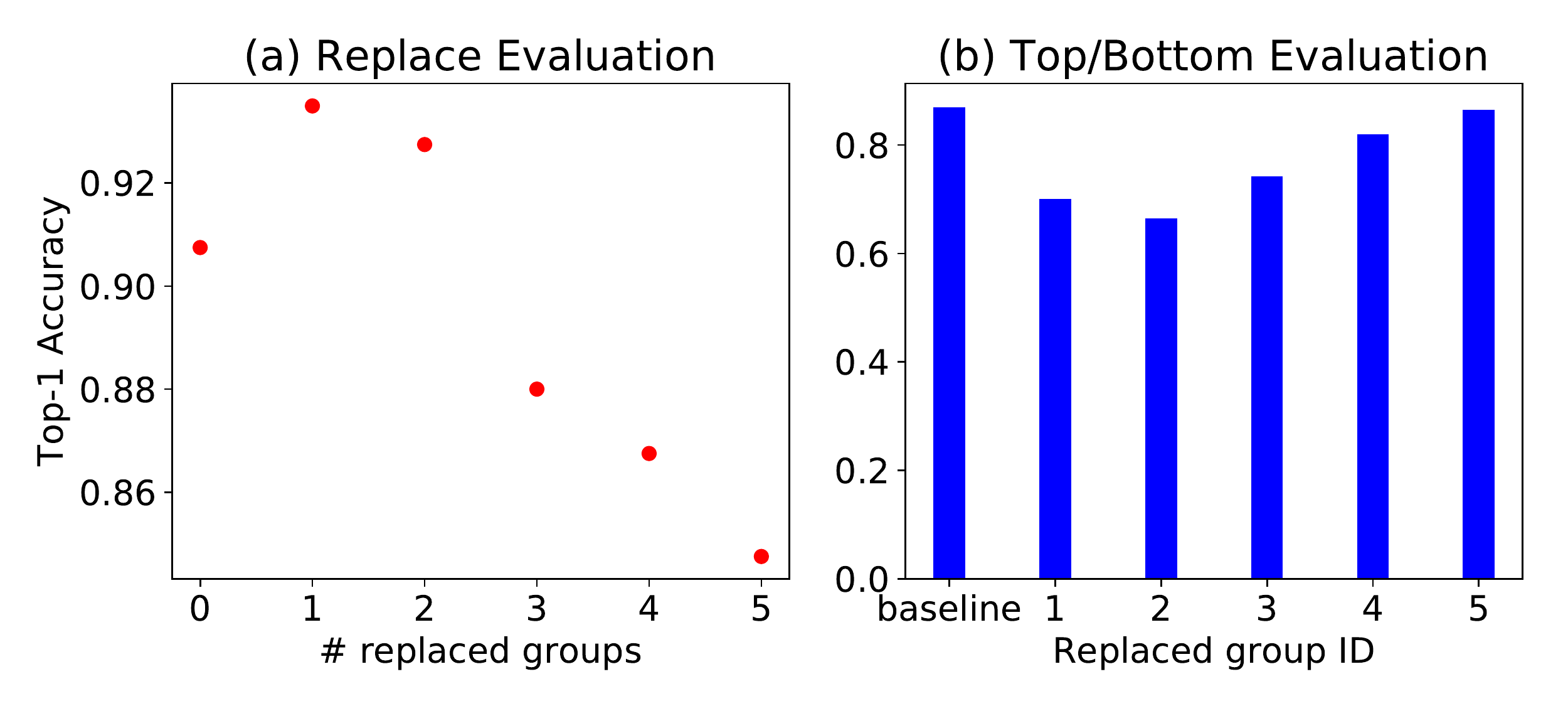}
\vspace{-1em}
\caption{(a) evaluates of accuracy by the number of replaced groups in VGG-16, and
(b) presents the impact of replacement positions.}
\label{fig:fine_tuning}
\end{figure}

We evaluate our method on a \emph{flowers} classification problem~\cite{flowers} where the original model is a pre-trained VGG-16.
The convolution layers in VGG-16 are replaced by their groups in this case.
Figure~\ref{fig:fine_tuning} presents the evaluation results in two aspects.
The left figure shows the final exploration results:
each point illustrates the accuracy and size of an explored model,
and the most efficient model is the one with the top convolution group replaced (the second point from left),
which is even better than the one with no layer replacement (the leftmost point).
Layers are consecutively replaced from top to bottom.
The right figure evaluates our assumption that
replacing the top convolution layers is more beneficial than the bottom ones.
In the figure,
a replaced group is closer to the top if its ID is bigger.
The rightmost column with the topmost group replaced achieves almost the same top-1 accuracy as the baseline model.
This evaluation is a proof-of-concept.
We will further evaluate this approach for other models and applications in future work.

\subsection{Final Toolflow}

Combining the model optimisation procedure described and the hardware optimisation and generation method in \secref{sec:design} and \secref{sec:design:fusion}, we can deduce the final toolflow for our framework as illustrated in Algorithm~\ref{alg:framework}. 
This algorithm can jointly explore the design space of CNN model and hardware
for efficient inference.


Given
$\mathcal{D}$ is the domain-specific dataset,
$\mathcal{R}$ are requirements,
$\mathcal{P}$ is platform specification,
and $\mathcal{M}$ are pre-trained models.
This algorithm is driven by the \textsc{ModelGen} procedure in line~\ref{alg:framework:next_model},
which can generate new models from pre-trained models and information from the current iteration,
such as the performance $p$ of the intermediate model $m$.
Basically, \textsc{DesignGen} automatically explores the hardware design space regarding $m$ and $\mathcal{P}$
and generates an optimised design $d$.
This design is then evaluated to get performance metrics $p$.
The best record is updated if the $p$ is better than the performance requirement $\mathcal{R}_{perf}$ and
the current best $p^*$.
The algorithm terminates when the accuracy is worse than the accuracy requirement $\mathcal{R}_{acc}$.
In case we have sufficient training budget,
we can loosen the terminating condition in line~\ref{alg:framework:term_cond}
by removing the accuracy requirement
and checking the accuracy until all possible models are searched.


\begin{algorithm}[!t]
\begin{footnotesize}
\caption{Pseudocode of the proposed framework.}
\label{alg:framework}
\begin{algorithmic}[1]
\Procedure{Framework}{$\mathcal{D}$, $\mathcal{R}$, $\mathcal{P}$, $\mathcal{M}$}
\State{$m \gets$ \Call{ModelGen}{$\mathcal{M}$}} \Comment{Initial model $m$}
\State{$m^*, p^*\gets m, 0$}\Comment{Initialise record}
\While{\Call{Valid}{$m$} $\wedge$ \Call{Acc}{$m$} $\ge \mathcal{R}_{acc}$}\label{alg:framework:term_cond}
\Comment{Check model accuracy}
\State{$d \gets$ \Call{DesignGen}{$m, \mathcal{P}$}} \Comment{Optimised design $d$}
\State{$p \gets$ \Call{Perf}{$d$, $m$, $\mathcal{P}$}} \Comment{Evaluate performance}
\If {$p \ge \mathcal{R}_{perf} \wedge p > p^*$}
\State $m^*, p^* \gets m, p$ \Comment{Update the best record}
\EndIf

\State{$m \gets$ \Call{ModelGen}{$\mathcal{M}$, $m$, $p$}} \Comment{Next model}\label{alg:framework:next_model}
\EndWhile
\EndProcedure
\end{algorithmic}
\end{footnotesize}
\end{algorithm}

\begin{table*}[!t]
\centering
\caption{Comparison of VGG-16 and ResNet-50 Performance with prior works}
\label{tab:eval_prev}
\begin{tabular}{|l|c|c|c|c|c|c|c|c|c|c|}
\hline

\multirow{2}{*}{}
& \multicolumn{6}{c|}{\textbf{VGG-16}}
& \multicolumn{4}{c|}{\textbf{ResNet-50}}
\\\cline{2-11}
& \cite{Qiu2016} & \cite{Zhang2016} & \cite{Ma2017a} & \cite{Lu2017} & \cite{Shen2018} & Ours
& \multicolumn{2}{c|}{\cite{Ma2017a}} & Ours (Plain) & Ours (Fused) 
\\\hline
              
\textbf{Year} & 2016 & 2016 & 2017 & 2017 & 2018 & 2018
              & 2017 & 2017 & 2018 & 2018\\\hline
            
\textbf{FPGA board }
& ZC706 & KU060  & GX1150 & ZCU102 & VCU440 & 5GSD8
& GXA7  & GX1150 & 5GSD8  & 5GSD8 \\\hline
              
\textbf{Tech.}& 28nm & 20nm & 20nm & 16nm & 16nm & 28nm 
              & 28nm & 20nm & 28nm & 28nm\\\hline
              
\textbf{Clock Freq. (MHz)}
    & 150 & 200 & 200 & 200 & 200 & 200
    & 150 & 200 & 200 & 200 \\\hline
\textbf{Bit width }
    & 16bit & 16bit & 16bit & 16bit & 16bit & 16bit 
    & 16bit & 16bit & 16bit & 16bit\\\hline
\textbf{Max DSP blocks}
    & 900 & 2760 & 1518 & 2520 & 2880 & 1963
    & 256 & 1518 & 1963 & 1963 \\\hline
\textbf{Perf. (GOPS)}
    & 137.0  & 266.0  & 720.2 & 2941 & 821.0 & 1928
    & 250.75 & 619.13 & 890.5 & 973.2  \\\hline

\end{tabular}
\end{table*}

\section{Evaluation}\label{sec:eval}


\begin{figure}[!t]
\centering
\includegraphics[width=0.48\textwidth]{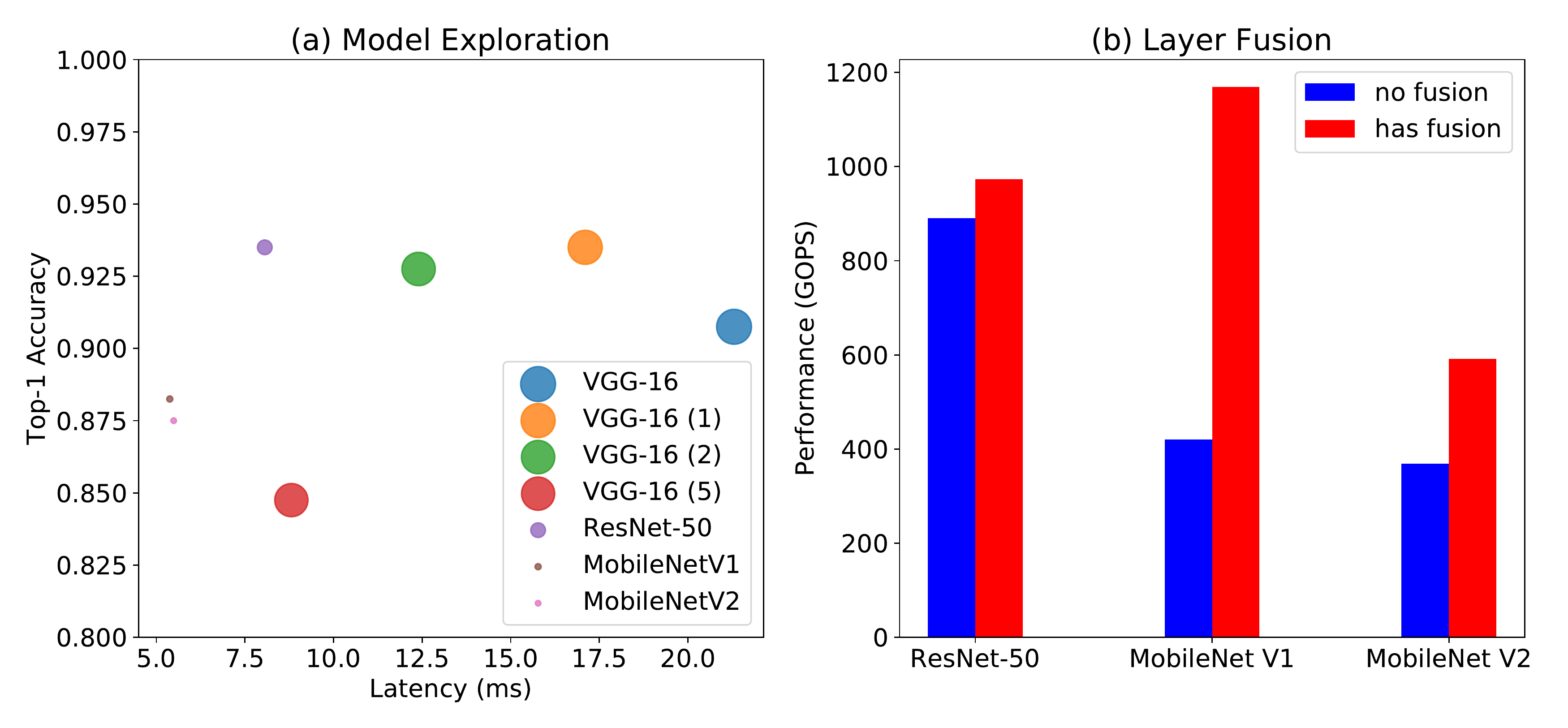}
\caption{
(a) visualises explored models and
(b) shows the improvement of layer fusion on efficient CNN models.
The size of a point in (a) shows its relative model size.}
\label{fig:join}
\vspace{-1em}
\end{figure}

In this evaluation, we look at the capability of TuRF by generating hardware design for typical efficient CNN models.
Then, we evaluate TuRF in terms of model transformation and optimisation by accepting conventional model VGG-16 pre-trained based on a large dataset and generating a set of smaller models with different number of groups replaced.
Finally, we compare our approach with previous work.

\subsection{Experimental Setup}
The domain-specific application that we select in this evaluation is the flowers classification 
problem~\cite{flowers}
mentioned above.
The data representation in the generated hardware designs is quantised to be 16 bit fixed-point,
which does not hurt the accuracy of the evaluated application.
All the CNN models evaluated here are built, trained and evaluated using the latest TensorFlow (v1.6).
Pre-trained models are downloaded directly from TF-Slim.
The experimental FPGA platform is Stratix-V 5SGSD8 on a Maxeler MPC-X node,
which contains 262.4K adaptive logic modules (ALM), 1963 variable-precision DSP blocks, and 2567 BRAM (M20K).
The bandwidth of off-chip data transfer is 38 GB/s.
The hardware template prototype is implemented in OpenSPL~\cite{openspl2013openspl}.
MaxCompiler (v2016.1.1) synthesises generated designs.

\subsection{Performance of Efficient CNN Models}

We first evaluate the performance of three popular efficient CNN models:
ResNet-50, MobileNet V1 and V2 generated by our framework and the results are shown in \tabref{tab:eval_result}.
Each model is fine-tuned to find the highest attainable top-1 accuracy
for flower classification.
From the table, ResNet-50 can achieve the best accuracy but it suffers from the worst performance in latency. The network size is also substantially larger when compared to the others.
On the other hand, MobileNet V1 is better than the V2 in terms of latency and accuracy with just a minor increase in network size.

We also study the benefits of layer fusion for convolution blocks by analysing the performance in GOPS. The layers within the convolution block are fused in each efficient model. \figref{fig:join} (b) compares performance, showing the
fused designs always outperform the implementations
without layer fusion.
Layer fusion is particularly effective for MobileNet V1,
revealing that depthwise separable blocks can be fused more effectively.
It also explains the compelling performance of MobileNet V1 as shown in \tabref{tab:eval_result} when compared to the V2.
Our framework allows users to choose which models to use, based on their requirements.
This involves a repeated execution of the exploration procedure and the model with higher satisfiability for
the given requirements will be chosen.

\subsection{Evaluation on Model Optimisation}
A pre-trained VGG-16 is used as an input to our framework
so as to evaluate its capability to perform model optimisation. 
The accuracy requirement supplied to the framework is gradually adjusted to generate implementations
with different number of groups replaced.
This enables us to understand the implications of replacing the standard convolution layer with various types of convolution block in conventional CNN model. 
\tabref{tab:eval_result} shows the results where VGG-16 (1), (2) and (5) imply one, two and five groups are replaced respectively. Essentially, VGG-16 (1) and (2) perform better than the original model in flowers classification regarding the accuracy and hardware efficiency.
The VGG-16 (5) only showcases minor performance gain with an enormous accuracy drop.

Furthermore, \figref{fig:join} (a) demonstrates the accuracy versus the latency and size
among all models.
MobileNet is more suitable for performance-aware applications while ResNet-50 and VGG-16 (2) are more appropriate for accuracy-aware applications.
Yet, the model size cannot drop significantly for VGG-16 because most parameters are occupied by
FC layers.


\subsection{Comparison with Previous Work}

To demonstrate the performance of our hardware template, we make a comparison to prior works related to automatic CNN accelerator generation on FPGA.
The original pre-trained CNN models, VGG-16 and ResNet-50,
are used in this experiment.
The convolution layer of our VGG-16 accelerator is not replaced by any convolution blocks to ensure a fair comparison.
The Winograd algorithm is applied to reduce the computation complexity.
Table~\ref{tab:eval_prev} shows that our approach is better than
most of the previous work and is still competitive with~\cite{Lu2017}
in the same technology.
Our performance normalised by 16 nm technology (3374 GOPS)
is higher than \cite{Lu2017} (2941 GOPS).
Moreover, to show that layer fusion can be beneficial for efficient convolution blocks,
we evaluate our accelerator on ResNet-50.
As shown in \tabref{tab:eval_prev}, 
our implementations outperform the ones given in~\cite{Ma2017a},
and the fused design can achieve the finest performance.
Here the \emph{plain} design is generated with only the Winograd algorithm,
and the \emph{fused} design performs layer fusion for all bottleneck blocks.


%

\section{Conclusion}\label{sec:conclusion}

This paper proposes TuRF, a new CNN optimisation framework inspired by efficient CNN architectures and transfer learning,
which supports domain-specific optimisations.
The novel aspects include a design template for various convolution blocks,
a layer fusion method,
and a model optimisation technique which allows layer replacement and fine-tuning of pre-trained CNNs.
The proposed approach is capable of producing some of the fastest CNN designs targeting FPGA implementations.
Further research includes design space exploration with functional evaluation tools, such as ADAM~\cite{ADAM2018}, and
extending our approach to support various applications.

\section*{Acknowledgements}
The support of the United Kingdom EPSRC (grant numbers EP/I012036/1, EP/L00058X/1, EP/L016796/1, EP/N031768/1 and EP/K034448/1), European Union Horizon 2020 Research and Innovation Programme (grant number 671653), Corerain, Intel, Maxeler and the Lee Family Scholarship is gratefully acknowledged.

\bibliographystyle{IEEEtran}
\bibliography{CNN,custom}
\end{document}